%% file: main_iarpa.tex
\documentclass[runningheads]{llncs}
\usepackage{graphicx}
\usepackage{amsmath,amssymb} 
\usepackage{color}
\DeclareMathOperator{\sign}{sign}
\graphicspath{{figs/}}
\usepackage[width=122mm,left=12mm,paperwidth=146mm,height=193mm,top=12mm,paperheight=217mm]{geometry}
\begin{document}
\pagestyle{headings}
\mainmatter

\title{Improving Network Robustness against Adversarial Attacks with Compact Convolution} 

\titlerunning{Improving Network Robustness against Adversarial Attacks with Compact Convolution}

\authorrunning{Ranjan~et~al.}

\author{Rajeev Ranjan, Swami Sankaranarayanan, Carlos D. Castillo, and Rama Chellappa}


\institute{Center for Automation Research, UMIACS,\\
	University of Maryland, College Park, MD 20742\\
	\email{ \{rranjan1,swamiviv,carlos,rama\}@umiacs.umd.edu}
}

\maketitle

\input{abstract}

\input{introduction}

\input{related}

\input{theory}


\input{pcn}

\input{experiments}

\input{supplementary}

\input{conclusion}

\section*{Acknowledgment}
This research is based upon work supported by the Office of the Director of National Intelligence (ODNI), Intelligence Advanced Research Projects
Activity (IARPA), via IARPA R\&D Contract No. 2014-14071600012. The views and conclusions contained herein are those of the authors and should
not be interpreted as necessarily representing the official policies or endorsements, either expressed or implied, of the ODNI, IARPA, or the U.S. Government. The U.S. Government is authorized to reproduce and distribute reprints for Governmental purposes notwithstanding any copyright annotation
thereon.

\clearpage

\bibliographystyle{splncs}
\bibliography{egbib}
\end{document}

%% file: abstract.tex
\begin{abstract}

Though Convolutional Neural Networks (CNNs) have surpassed human-level performance on tasks such as object classification and face verification, they can easily be fooled by adversarial attacks. These attacks add a small perturbation to the input image that causes the network to mis-classify the sample. In this paper, we focus on neutralizing adversarial attacks by compact feature learning. In particular, we show that learning features in a closed and bounded space improves the robustness of the network. We explore the effect of L2-Softmax Loss, that enforces \textit{compactness} in the learned features, thus resulting in enhanced robustness to adversarial perturbations. Additionally, we propose compact convolution, a novel method of convolution that when incorporated in conventional CNNs improves their robustness. Compact convolution ensures feature compactness at every layer such that they are bounded and close to each other. Extensive experiments show that Compact Convolutional Networks (CCNs) neutralize multiple types of attacks, and perform better than existing methods in defending adversarial attacks, without incurring any additional training overhead compared to CNNs. 
   
\keywords{Adversarial Attacks, Convolutional Neural Networks, Compact Convolution, Adversarial Robustness}

\end{abstract}



%% file: introduction.tex
\section{Introduction}

In recent years, CNNs have gained tremendous popularity because of their impressive performance on many vision-related tasks. They are being widely used in many practical applications such as self-driving cars, face verification, etc. However, it has been shown that CNNs are vulnerable to small adversarial perturbations which, when added to the input image, can cause the network to mis-classify with high confidence~\cite{szegedy2013intriguing,goodfellow2014explaining,moosavi2016deepfool,moosavi2016universal}. Adversarial images thus generated are often visually indistinguishable from the original images.

\begin{figure}[t]
	\begin{center}
		\includegraphics[width=9cm, height=5.2cm]{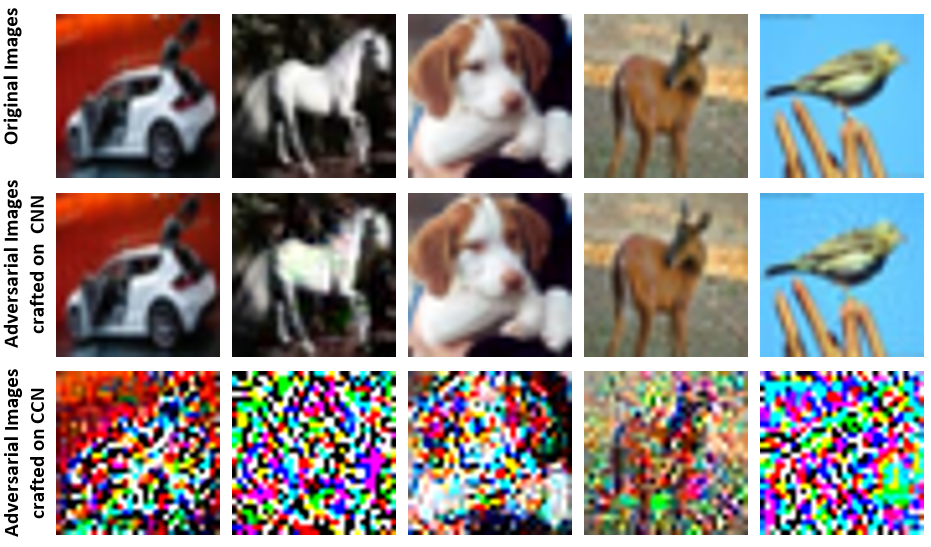}
	\end{center}
	\caption{Five test samples from CIFAR10~\cite{krizhevsky2009learning} dataset~(top) that are correctly classified by both CNN and CCN. Corresponding adversarial images crafted using DeepFool~\cite{moosavi2016deepfool} attack and misclassified by CNN~(middle) and CCN~(bottom). The adversarial attacks on CCN can be detected trivially by human observer.}
	\label{fig:sample}
\vspace*{-4mm}
\end{figure}

Adversarial attacks have emerged as a potential threat to CNN-based systems. Adversarial images can be used by a suspect to fool a face verification system, by letting the person go unidentified. These attacks can also cause self-driving cars to mis-classify scene objects such as a stop sign leading to adverse effects when these systems are deployed in real time. As networks move from the research labs to the field, they need to be designed in a way that they are not only accurate, but also robust to adversarial perturbations. Several recent works have been proposed to improve robustness, such as adversarial training~\cite{goodfellow2014explaining,sankaranarayanan2017regularizing}, gradient masking~\cite{tramer2017ensemble}, etc. In this work, we impose constraints on the sensitivity of the learned feature space and using our insight, propose a modified convolution operation that can desensitize the learned mapping in the direction of adversarial perturbations.

It has been hypothesized that CNNs learn a highly non-linear manifold on which the images of same class lie together, while images from different class are separated. Hence, the original image and the adversarial image lying close to each other in Euclidean space, are far separated on the manifold or in feature space. When designing a robust classifier, we would like to address the following question: \textit{Can we bring the original and perturbed images closer in the feature space of a learned mapping to improve its robustness?} To address this question, we employ the property of \textit{compactness} in the context of feature learning that would enhance a network's robustness to adversarial attacks. \textit{Compactness} enforces the features to be bounded  and lie in a closed space. It reduces the degree of freedom for the features to be learned. This restricts the extent to which a feature for perturbed image can move, making it less likely to cross the class boundary.

To enforce \textit{compactness} in the feature space, we explore the $L_{2}$-Softmax Loss proposed by Ranjan~et~al.~\cite{ranjan2017l2}. The $L_{2}$-Softmax Loss establishes \textit{compactness} by constraining the features to lie on a hypersphere of fixed radius, before applying the softmax loss. It brings the intra-class features close to each other and separates the inter-class features far apart. In this way, features from the original and the adversarial image are closer to each other using $L_{2}$-Softmax Loss, compared to training with regular softmax loss (see Fig.~\ref{fig:loss_functions}).

Using these insights, we propose a novel convolution method, called \textit{compact convolution}, that significantly enhances a network's robustness by ensuring compact feature learning at every layer of the network. A compact convolution module applies the $L_{2}$-normalization step and scaling operations to every input patch before applying the convolutional kernel in a sliding window fashion. Compact Convolutional Networks (CCNs), built using these modules, are highly robust compared to a typical CNN. Fig.~\ref{fig:sample} shows some sample images and corresponding adversarial attacks generated using DeepFool~\cite{moosavi2016deepfool} to fool a CNN and a CCN. The adversarial samples for CCN can easily be distinguished from the original samples by a human observer. The figure shows that CCNs are robust to small adversarial perturbations such that to fool a CCN the magnitude of perturbations required is much higher, which completely distorts the image and can be detected easily. The paper makes the following key contributions:

\noindent
$\bullet$ We explore the property of  \textit{Compactness} in the context of feature learning, and demonstrate its effectiveness in making a network robust to adversarial attacks.\\
$\bullet$ We propose to use L2-Softmax Loss as a defense mechanism against adversarial perturbations.\\
$\bullet$ We propose compact convolutional modules that increases the network stability by enforcing \textit{Compactness} to features at every layer in the network.\\
$\bullet$ We achieve new state-of-the-art results on defending white-box as well as black-box attacks.

%% file: related.tex
\section{Related Works}
\label{sec:related}

A lot of research has gone into generating adversarial perturbations to fool a deep network.  Szegedy~et~al.~\cite{szegedy2013intriguing} first showed the existence of adversarial perturbations in CNNs and proposed a L-BFGS based optimization scheme to generate the same. Later, Goodfellow~et~al.~\cite{goodfellow2014explaining} proposed Fast Gradient Sign Method~(FGSM) to generate adversarial samples. DeepFool~\cite{moosavi2016deepfool} attack iteratively finds a minimal perturbation required to cause a network to mis-classify. Other recently proposed adversarial attacks include Jacobian-based Saliency Map Approach (JSMA)~\cite{papernot2016limitations}, Carlini-Wagner~(CW) attack~\cite{carlini2017towards}, Universal Perturbations~\cite{moosavi2016universal}, etc.

To safeguard the network from adversarial attacks, researchers have focused on two approaches: 1) \textit{Adversarial Detection}, and 2) \textit{Adversarial Defense}. Methods based on adversarial detection~\cite{lu2017safetynet,metzen2017detecting,grosse2017statistical,feinman2017detecting} attempt to detect an adversarial sample before passing it through the network for inference. These methods put an extra effort in designing a separate adversarial detector which itself has the risk of being fooled by the attacker. Recently, Carlini and Wagner~\cite{carlini2017adversarial} showed that most of the adversarial detectors are ineffective and can be fooled.

The methods based on adversarial defense aim at improving the network's robustness to classify adversarial samples correctly. One way to achieve robustness is by simultaneously training the network with clean and adversarial samples~\cite{goodfellow2014explaining,miyato2017virtual,shaham2015understanding,kurakin2016adversarial}. These methods are stable to the attack on which they are trained, but ineffective against a different attack. Preprocessing the input to nullify the adversarial effect is another way to defend the network~\cite{dziugaite2016study,guo2017countering}. Few methods have focused on modifying the network topology or optimization procedure for adversarial defense. Gu and Rigazio~\cite{gu2014towards} proposed  Deep Contractive Network that adds a smoothness penalty on the partial derivatives at every layer. Cisse~et~al.~\cite{cisse2017parseval} proposed Parseval Networks that improves robustness by enforcing the weight matrices of convolutional and linear layers to be Parseval tight frames. Papernot~et~al.~\cite{papernot2016distillation} showed that knowledge distillation with high temperature parameter can be used as defense against adversarial samples. Warde~et~al.~\cite{warde201611} showed that a similar robustness as defensive distillation can be obtained by training the network with smooth labels. Zantedeschi~et~al.~\cite{zantedeschi2017efficient} used Bounded ReLU activations to enhance network's stable to adversarial perturbations. 

While these methods have focused on improving defense to adversarial attacks in general, most of them have focused on white box attacks and incur additional computational overhead during training. In this work, we propose an approach to achieve adversarial defense in CNNs using compact convolutions which can be seamlessly integrated into any existing deep network architecture. We further demonstrate its effectiveness against both white box and black box adversarial attacks.

%% file: theory.tex
\section{Compact Learning for Adversarial Defense}
\label{sec:adv_defense}


\subsection{Compactness}
\label{sec:compactness}

\textit{Compactness} is a property associated with a subset of Euclidean space which is closed and bounded. A space is closed when it contains all its limiting points. A space is bounded when all its points lie within a fixed distance of each other. Euclidean space in itself is not \textit{compact}, since it is not bounded.

The features obtained from a typical CNN are not \textit{compact}, since the softmax loss does not constrain them to lie in a closed or bounded space. It distributes the features in the Euclidean space such that the overall training loss is minimized. In a way, it over-fits to the training domain. Thus an adversarially perturbed image, although close to the original image in input space, can lie very far away in the Euclidean feature space. On the other hand, features learned on a compact space have restricted degrees of freedom which does not allow the adversarial features to move far away from the original features. A given perturbed image would lie farther from the original image in the Euclidean space compared to a compact space. Thus compact feature learning helps in improving the robustness of the network.

One way to enforce compactness on the CNN features is to restrict them to lie on a compact space during training.  A simple example of a compact space is a hypersphere manifold, which is both closed and bounded. $L_{2}$-Softmax Loss~\cite{ranjan2017l2}~(discussed in Section~\ref{sec:L2SM}), performs compact feature learning by constraining the features to lie on a hypersphere of a fixed radius.

\subsection{$L_{2}$-Softmax Loss}
\label{sec:L2SM}
$L_{2}$-Softmax loss was recently proposed by Ranjan~et~al.~\cite{ranjan2017l2} for improving the task of face verification. The loss imposes a constraint on the deep features to lie on a hypersphere of a fixed radius, before applying the softmax penalty. The loss is defined as:

\begin{equation}
\small
\label{eq:l2smloss}
L_{S} = -\sum_{i=1}^{m} \log \frac{e^{W_{y_{i}}^{T}(\frac{\alpha \mathbf{x}_{i}}{\|\mathbf{x}_{i}\|_{2}}) + b_{y_{i}}}}{\sum_{j=1}^{n} e^{W_{j}^{T}(\frac{\alpha \mathbf{x}_{i}}{\|\mathbf{x}_{i}\|_{2}}) + b_{j}}},
\end{equation}

where $\mathbf{x}_{i}$ is the $i^{th}$ deep feature for the class label $y_{i}$, $W_{j}$ is the weight and $b_{j}$ is the bias corresponding to the class $j$, $\alpha$ is a positive scalar parameter, and $m$, $n$ are the batch-size and number of classes respectively. The features are first normalized to unit length and then scaled by $\alpha$ before passing it through the softmax classifier. Constraining the features to lie on a hypersphere reduces the intra-class variations and enhances the inter-class separability.

\begin{figure*}[htp!]
 \centering
\includegraphics[width=5.5cm]{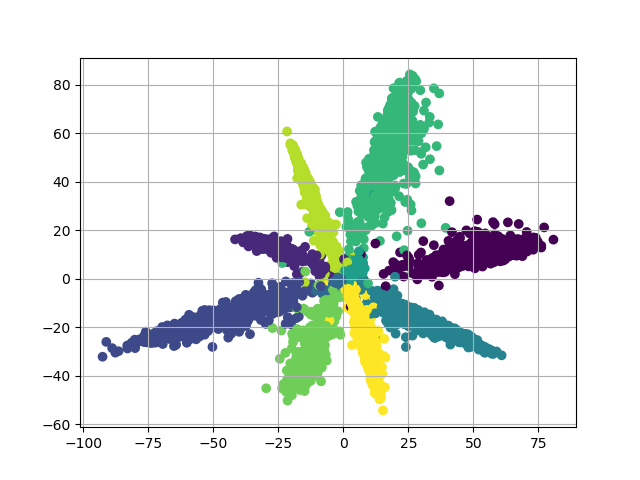}\hskip5pt\includegraphics[width=5.5cm]
{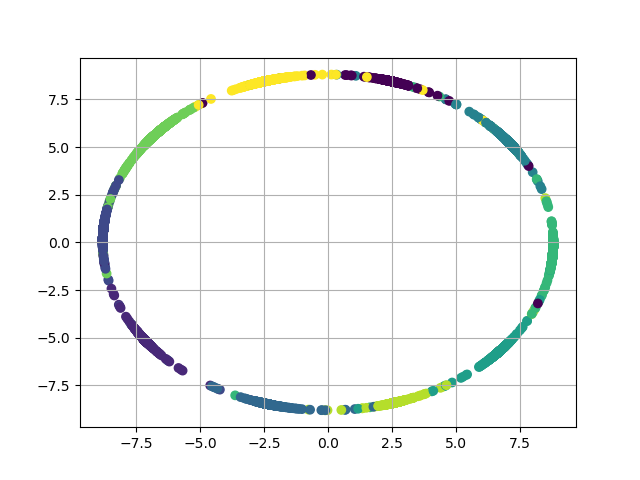}\\
(a)\hskip150pt(c)\\
\caption{Feature vizualization from the last layer (dimension=$2$) of CNN trained on MNIST~\cite{lecun2010mnist} with (a) Softmax Loss, (b) $L_{2}$-Softmax Loss~\cite{ranjan2017l2}.}
\label{fig:loss_functions}
\vspace*{-4mm}
\end{figure*} 

From Fig.~\ref{fig:loss_functions}(a), we can see that the features trained using softmax loss do not satisfy the \textit{compactness} property since the feature space is not closed or bounded. The $L_{2}$-Softmax Loss constrains the features to lie on a hypersphere manifold which is a closed space. Also, the $L_{2}$-norm of the feature vectors is always constant and equal to $\alpha$, which makes the feature space bounded. Hence, the $L_{2}$-Softmax Loss obeys the \textit{compactness} property (Fig.~\ref{fig:loss_functions}(b)). Experimental analysis (see section~\ref{sec:experiments}) shows that $L_{2}$-Softmax Loss is more robust to adversarial attacks than softmax loss. To the best of our knowledge, $L_{2}$-Softmax loss has not been used for adversarial defense before.

\subsection{Sub-linearity of Adversarial Examples with Compact Learning}
In this section, we show that compact feature learning restricts the change in activation due to perturbation $\eta$ to grow sub-linearly with the input dimension. Consider a linear model with weight vector $w$ and input vector $x$ with dimension $n$. Let the adversarial example be represented as $\tilde{x}$ = $x$ + $\eta$. The activation is given by the dot product:
\begin{equation}
\label{eq:accstegano}
w^{T}\tilde{x} = w^{T}x + w^{T}\eta 
\end{equation}
It was shown by Goodfellow et al.~\cite{goodfellow2014explaining} that the change in activation $w^{T}\eta$ will grow with $\epsilon m n$, where $m$ is the average magnitude of the weight vector, and $\|\eta\|_{\infty} < \epsilon$. Thus, the effect of adversarial perturbation increases linearly with the input dimension. Now, with compact feature learning using $L_{2}$-Softmax loss with $\alpha=1$, (\ref{eq:accstegano}) can be rewritten as:

\begin{equation}
\label{eq:accstegano_l2s}
w^{T}\frac{\tilde{x}}{\|\tilde{x}\|_{2}} = w^{T}\frac{x}{\|x+\eta\|_{2}} + w^{T}\frac{\eta}{\|x+\eta\|_{2}}.
\end{equation}







The second term in~(\ref{eq:accstegano_l2s}) is the change in activation due to adversarial perturbation. The denominator in this term, which is the $L_2$-norm of the perturbed signal ($\|x+\eta\|_{2}$), changes proportional to the square root of the dimension of $x$. This can be intuitively thought of as follows. Let the average absolute activation of $x$ be $p$. Then the $L_2$-norm of the signal $x$ can be well approximated by $p~\sqrt[]{n}$. While the scaling factor ($p$) depends on the actual variations in the feature value, the dependency on the dimension is clear. Applying this insight to (~\ref{eq:accstegano_l2s}), we see that the change in the activation now grows at the rate of $~\sqrt[]{n}$ instead of $n$, due to the normalizing factor in the denominator. Thus, for a given input dimension, the change in activation due to adversarial perturbation would be smaller for compact features compared to the typical CNN features, which makes it ideal for training a robust network.


%% file: pcn.tex
\section{Compact Convolution}
\label{sec:pcn}

\begin{figure}[htp!]
\centering
\includegraphics[width=7.5cm, height=4.0cm]{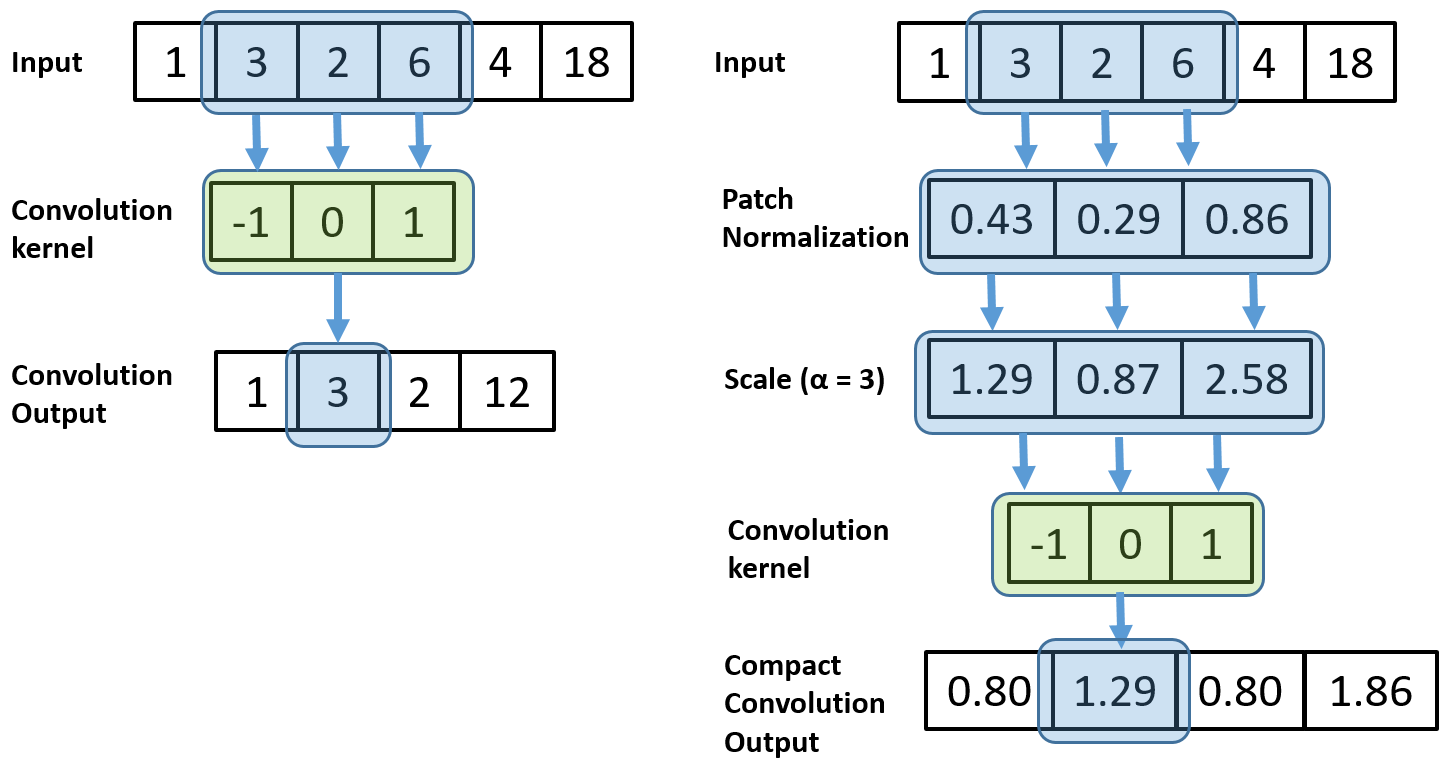}
\caption{A toy example for convolution (left) and compact convolution (right).}
\label{fig:pconv_toy}
\vspace*{-4mm}
\end{figure}

The $L_{2}$-Softmax loss~\cite{ranjan2017l2} enforces \textit{compactness} only to deep features from the last layer of CNN. It was motivated by efficient representation of normalized feature at the output space, whereas in this paper we want to reduce the sensitivity of the activations at each layer to spurious perturbations. Hence, we propose to extend the \textit{compactness} property to features from the intermediate layers of CNNs as well. A typical CNN is a hierarchy of convolutional and fully-connected layers stacked with non-linear activation functions after every layer. A discrete convolution is a linear function applied on a patch of a signal in a sliding window fashion. Let $W$ be the convolution kernel of size $2k+1$, $\mathbf{x}_{n,k}$ be an input patch defined as:

\begin{equation}
\small
\label{eq:input_patch}
\mathbf{x}_{n,k} = [x(n-k), x(n-k+1), ..., x(n+k)],
\end{equation}

where $x(n)$ is the $n^{th}$ element of input vector $\mathbf{x}$. The convolution operation is represented as:

\begin{equation}
\small
\label{eq:conv}
y(n) = W^{T} \mathbf{x}_{n,k},
\end{equation}

where $y(n)$ is the $n^{th}$ element of the output vector $\mathbf{y}$. To enforce \textit{compactness} in convolutional layers, we need to ensure that every input patch at a given location is first $L_{2}$-normalized and scaled before multiplying with the convolutional kernel $W$. Formally, we want the convolution output ($\tilde{y}(n)$) at position $n$ to be:

\begin{equation}
\small
\label{eq:pconv}
\tilde{y}(n) = W^{T} \frac{(\alpha \mathbf{x}_{n,k})}{\|\mathbf{x}_{n,k}\|_{2}+\delta},
\end{equation}

where $\delta$ is a small constant added to avoid division by zero. We call this new method of patch-normalized convolution as \textit{compact convolution}. A toy example depicting the difference between typical convolution and compact convolution is shown in Fig.~\ref{fig:pconv_toy}. 

\begin{figure}[htp!]
\centering
\includegraphics[width=12cm, height=2.7cm]{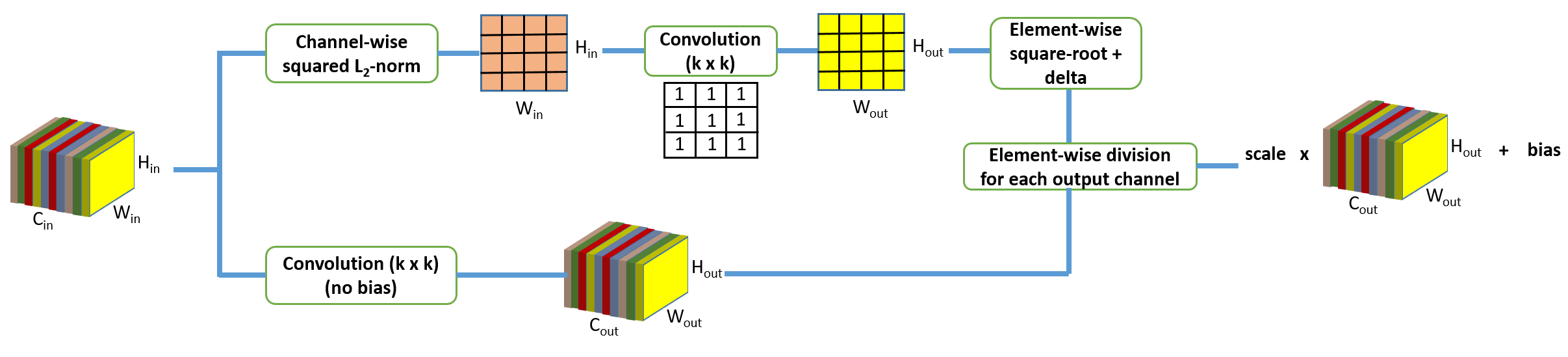}
\caption{Block diagram of a compact convolution module.}
\label{fig:pconv_module}
\vspace*{-4mm}
\end{figure}

In a deep CNN, the convolution kernel is typically applied to high dimensional feature maps. Normalizing every feature patch before multiplying with the convolutional kernel is computationally expensive and redundant, since the patches are overlapping. To implement compact convolution efficiently in a deep network, we propose a compact convolution module (shown in Fig.~\ref{fig:pconv_module}). We split the input feature map into two branches. The first branch carries out the traditional convolution operation with parameters of size $k \times k$, without bias addition. The second branch first computes the sum of squares along the channel dimension of the input. Subsequently, it is convolved with a $k \times k$ kernel containing fixed value of all ones. This step provides the squared $L_{2}$-Norm of sliding-window patches for every output location in a feature map. We perform element-wise square-root on top of it and add a small constant $\delta=0.01$. Lastly, each channel of the convolutional output from the first branch is divided element-wise with the output from the second branch. We then scale the final output with a learnable scalar parameter $\alpha$ and add the bias term. The compact convolution module uses just one extra learnable parameter ($\alpha$) compared to the traditional convolutional layer. 

Since the linear operation in fully-connected layers is a special case of convolution, the compact convolution operation for these layers results in $L_{2}$-normalization and scaling of the feature vectors. It constrains the features to lie on a hypersphere of fixed radius ($\alpha$), before applying the dot product with the layer parameters. We call the deep networks with compact convolution modules as Compact Convolutional Networks (CCNs). They follow the \textit{compactness} property at every layer of the network, which greatly enhances their robustness against multiple kinds of adversarial attacks. The $L_{2}$-Softmax Loss~\cite{ranjan2017l2} inherently gets applied in CCNs.


%% file: experiments.tex
\section{Experiments}
\label{sec:experiments}

We evaluate the effectiveness of the proposed defense methods on MNIST~\cite{lecun2010mnist}, CIFAR10~\cite{krizhevsky2009learning} and ImageNet~\cite{russakovsky2015imagenet} datasets. The MNIST~\cite{lecun2010mnist} dataset contains $60,000$ training and $10,000$ test images of handwritten digits. The images are $28 \times 28$ dimensional with values in $[0, 1]$. CIFAR10~\cite{krizhevsky2009learning} dataset contains $50,000$ training and $10,000$ test images of $10$ classes. The images are $32 \times 32 \times 3$ dimensional, with values scaled between $0$ and $1$.

We use two well-known methods for crafting adversarial attacks: Fast Gradient Sign Method~\cite{goodfellow2014explaining} (FGSM) and DeepFool~\cite{moosavi2016deepfool}. The FGSM attack adds the sign of the gradient to the input, scaled by the factor $\epsilon$ as shown in~(\ref{eq:fgsm})

\begin{equation}
\small
\label{eq:fgsm}
\mathbf{\tilde{x}} = \mathbf{x} + \epsilon \sign(\nabla_{\mathbf{x}} J(\mathbf{\theta}, \mathbf{x}, y)),
\end{equation}

where $\mathbf{x}$ is the input image, $\mathbf{\tilde{x}}$ is the crafted adversarial image, $\nabla_{\mathbf{x}} J(\mathbf{\theta}, \mathbf{x}, y))$ is the gradient of the cost function with respect to the input. This method is very fast, since it uses a single backward step to generate the attack. On the other hand, DeepFool~\cite{moosavi2016deepfool} iteratively finds the minimal perturbation required to mis-classify the input in the direction of the nearest class boundary. Though slower than FGSM~\cite{goodfellow2014explaining}, DeepFool~\cite{moosavi2016deepfool} can generate adversarial images with smaller magnitude of perturbations, which are indistinguishable to human observer. We use the Foolbox~\cite{rauber2017foolbox} library to generate these attacks.

The network architectures used for training are provided in the supplementary material. For training on MNIST~\cite{lecun2010mnist}, we use the architecture proposed by Papernot~et~al.~\cite{papernot2016distillation}. The learning rate is set to $0.1$ for the first thirty epochs, and decreased by a factor of $0.1$ after every ten epochs. We train the network for fifty epochs. For training on CIFAR10~\cite{krizhevsky2009learning}, we use the standard VGG11~\cite{simonyan2014very} network. The convolutional layers use $3 \times 3$ kernels with padding of one. We start with a learning rate of $0.1$ which is decreased by a factor of $0.2$ after $60$, $120$ and $160$ epochs. We train the network for $200$ epochs. We use SGD with momentum ($0.9$) and weight decay ($5 \times 10^{-4}$) for all our training. We use mean subtraction of $0.5$ as a pre-processing step.


We compare and evaluate the following defense methods against adversarial attacks:\\
$\bullet$ \textbf{SM} - The baseline model trained using the typical Softmax loss function.\\
$\bullet$ \textbf{LS} (\textit{Label Smoothing}) - The model trained using Softmax Loss with soft labels \{$\frac{1}{90}$,$0.9$\}~\cite{warde201611} instead of discrete labels \{$0$,$1$\}\\
$\bullet$ \textbf{BReLU} (\textit{Bounded ReLU}) - The model trained using bounded ReLU~\cite{zantedeschi2017efficient} activation function instead of ReLU. The activations are clipped between $[0,1]$.\\
$\bullet$ \textbf{L2SM} ($L_{2}$-\textit{Softmax}) - The model trained using $L_{2}$-Softmax Loss~\cite{ranjan2017l2} as discussed in Section~\ref{sec:L2SM}.\\
$\bullet$ \textbf{CCN} (\textit{Compact Convolutional Network}) - The model trained with compact convolution modules instead of traditional convolutional and fully-connected layers, as discussed in Section~\ref{sec:pcn}.\\
$\bullet$ \textbf{CCN+LS} (\textit{Compact Convolutional Network} with \textit{Label Smoothing}) - A CCN trained using soft labels \{$\frac{1}{90}$,$0.9$\}
\\

The experiments are organized as follows. Section~\ref{sec:wb_attack} evaluates the proposed models on MNIST~\cite{lecun2010mnist}, CIFAR10~\cite{krizhevsky2009learning} and ImageNet~\cite{russakovsky2015imagenet} datasets, against FGSM and DeepFool attacks in a white-box setting. It also analyzes the effect of adversarial training on different models. Section~\ref{sec:bb_attack} evaluates the robustness in a black-box setting, and discusses the transferability of various attacks. Section~\ref{sec:norm} compares the robustness using other feature normalization methods such as Local Response Normalization~\cite{krizhevsky2012imagenet} and batch-normalization~\cite{ioffe2015batch}.

\subsection{White-Box Attacks}
\label{sec:wb_attack}
In a white-box attack, the attacker has full access to the network to be attacked. For each of the defense methods, we generate FGSM~\cite{goodfellow2014explaining} and DeepFool~\cite{moosavi2016deepfool} attack for MNIST~\cite{lecun2010mnist} and CIFAR10~\cite{krizhevsky2009learning} testset. 

Table~\ref{tbl:wb_fgsm} provides the classification accuracy of various defense methods on adversarial examples crafted using FGSM~\cite{goodfellow2014explaining} for MNIST~\cite{lecun2010mnist} and CIFAR10~\cite{krizhevsky2009learning} testset. We perform evaluations for four different $\epsilon$ values \{$0.1$,$0.2$,$0.3$,$0.4$\}. Higher values of $\epsilon$ lead to larger perturbation, thus decreasing accuracy. Note that $\epsilon = 0$ corresponds to the clean test samples without any adversarial perturbation. From the table, we find that both \textit{CCN} and \textit{CCN+LS} are highly robust to FGSM attack, with minimal degradation in accuracy. Specifically, for $\epsilon = 0.3$, \textit{CCN} achieves an accuracy of $81.38\%$ on MNIST~\cite{lecun2010mnist} which is more than $2\times$ factor improvement over the baseline model with accuracy $31.76\%$. Label Smoothing along with CCN (\textit{CCN+LS}) further enhances the robustness to achieve an accuracy of $89.73\%$.  The \textit{L2SM} model shows significant improvement over the baseline, which establishes its robustness. A similar trend is observed with FGSM~\cite{goodfellow2014explaining} attack on CIFAR10~\cite{krizhevsky2009learning} testset. Since CIFAR10 is a harder dataset, we use the $\epsilon$ values of \{$\frac{2}{255}$,$\frac{4}{255}$,$\frac{8}{255}$,$\frac{16}{255}$\} to craft the FGSM attack. For $\epsilon = \frac{8}{255}$, we observe $5\times$ improvement using \textit{CCN} and \textit{CCN+LS}, compared to the baseline model.

\begin{table}[htp!]
\caption{Accuracy ($\%$) on MNIST~\cite{lecun2010mnist} and CIFAR10~\cite{krizhevsky2009learning} for FGSM~\cite{goodfellow2014explaining} attack. The best accuracy is shown in bold and the second-best accuracy is underlined}
\label{tbl:wb_fgsm}
\begin{center}
\tabcolsep=0.10cm
\scalebox{0.875}{
\begin{tabular}{|c||c|c|c|c|c||c|c|c|c|c|}
\hline
 & \multicolumn{5}{c|}{MNIST} & \multicolumn{5}{c|}{CIFAR10}\\
\hline
Method & $\epsilon$=0 & $\epsilon$=0.1 & $\epsilon$=0.2 & $\epsilon$=0.3 & $\epsilon$=0.4 & $\epsilon$=0 & $\epsilon$=2 & $\epsilon$=4 & $\epsilon$=8 & $\epsilon$=16 \\
\hline\hline
SM & 99.50 & 92.61 & 63.71 & 31.76 & 19.39 & \underline{91.14} & 60.82 & 34.62 & 14.30 & 8.57\\
LS~\cite{warde201611} & \underline{99.53} & 95.62 & 85.13 & 44.25 & 15.57 & 91.03 & 66.7 & 58.4 & 54.07 & 51.05\\
BReLU~\cite{zantedeschi2017efficient} & \textbf{99.54} & 95.29 & 74.83 & 42.51 & 19.41 & 90.87 & 61.49 & 35.02 & 18.01 & 13.73\\
\hline
L2SM & 99.48 & 94.51 & 79.68 & 60.32 & 45.16 & \textbf{91.37} & 70.58 & 65.39 & 63.71 & 62.59\\
CCN & 99.50 & \underline{96.99} & \underline{91.64} & \underline{81.38} & \underline{60.65} & 90.54 & \underline{73.72} & \underline{71.47} & \underline{69.65} & \underline{66.23}\\
CCN+LS & 99.43 & \textbf{97.26} & \textbf{93.68} & \textbf{89.73} & \textbf{75.95} & 90.51 & \textbf{80.93} & \textbf{79.2} & \textbf{76.8} & \textbf{71.38}\\
\hline
\end{tabular}
}
\end{center}
\vspace*{-4mm}
\end{table}

Table~\ref{tbl:wb_deepfool} provides the classification accuracies of different defense methods against DeepFool~\cite{moosavi2016deepfool} attack, on MNIST and CIFAR10 datasets. Since, DeepFool is an iterative attack, it will mostly find a perturbation to fool the network. To evaluate using DeepFool, the iterations are carried out until the adversarial perturbation ($\mathbf{\eta}$) causes the network to mis-classify, or the ratio of $L_{2}$-norm of the perturbation ($\mathbf{\eta}$) and the input ($\mathbf{x}$) reaches the \textit{max-residue-ratio} ($d$), as given in~(\ref{eq:res-ratio}).
\begin{equation}
\small
\label{eq:res-ratio}
\frac{\|\mathbf{\eta}\|_{2}}{\|\mathbf{x}\|_{2}} \leq d
\end{equation}

The evaluations are carried out with different values of $d$ = \{$0.1$,$0.2$,$0.3$,$0.4$,$0.5$\}. From Table~\ref{tbl:wb_deepfool}, we find that \textit{LS} and \textit{CCN} perform comparably against DeepFool attack on MNIST. The model \textit{CCN+LS} achieves the best performance since it leverages the qualities of both \textit{LS} and \textit{CCN} models.
On CIFAR10 dataset, \textit{CCN} performs better than \textit{LS} against DeepFool attack. The effect of label smoothing is less for CIFAR10, since the probability scores are lower owing to the difficulty of the dataset. The models \textit{CCN} and \textit{CCN+LS} significantly outperform other defense methods against DeepFool attack.

\begin{table}[htp!]
\caption{Accuracy ($\%$) for DeepFool~\cite{moosavi2016deepfool} attack (with different max-residue-ratio $d$) on MNIST~\cite{lecun2010mnist} and CIFAR10~\cite{krizhevsky2009learning} testset. The best accuracy is shown in bold and the second-best accuracy is underlined}
\label{tbl:wb_deepfool}
\begin{center}
\tabcolsep=0.10cm
\scalebox{0.875}{
\begin{tabular}{|c||c|c|c|c|c||c|c|c|c|c|}
\hline
 & \multicolumn{5}{c|}{MNIST} & \multicolumn{5}{c|}{CIFAR10}\\
\hline
Method & $d$=0.1 & $d$=0.2 & $d$=0.3 & $d$=0.4 & $d$=0.5 & $d$=0.1 & $d$=0.2 & $d$=0.3 & $d$=0.4 & $d$=0.5\\
\hline\hline
SM & 93.24 & 63.22 & 18.73 & 1.89 & 0.53 & 5.73 & 5.72 & 5.72 & 5.72 & 5.72\\
LS~\cite{warde201611} & 96.85 & \underline{92.2} & \underline{88.12} & \underline{82.05} & \underline{70.08} & 57.33 & 43.93 & 31.77 & 21.89 & 14.6\\
BReLU~\cite{zantedeschi2017efficient} & 96.15 & 84.33 & 70.13 & 57.25 & 45.5 & 22.44 & 15.49 & 10.48 & 7.29 & 6.37\\
\hline
L2SM & 96.5 & 88.22 & 75.94 & 59.47 & 40.31 & 66.63 & 56.93 & 47.89 & 38.52 & 28.85\\
CCN & \underline{97.06} & 92.18 & 83.46 & 69.25 & 48.54 & \underline{73.17} & \underline{71.78} & \underline{70.24} & \underline{68.13} & \underline{64.02}\\
CCN+LS & \textbf{97.76} & \textbf{95.42} & \textbf{92.43} & \textbf{88.65} & \textbf{84.01} & \textbf{80.8} & \textbf{79.38} & \textbf{78.08} & \textbf{76.43} & \textbf{74.22}\\
\hline
\end{tabular}
}
\end{center}
\vspace*{-4mm}
\end{table}

We also provide a quantitative measure of model's robustness against DeepFool~\cite{moosavi2016deepfool} attack in Table~\ref{tbl:robustness}. The robustness $\rho_{adv}(\hat{k})$ is computed using~(\ref{eq:robustness}). 
\begin{equation}
\small
\label{eq:robustness}
\rho_{adv}(\hat{k}) = \mathbb{E}_{\mathbf{x}} \frac{\|\mathbf{r}_{\mathbf{x},\hat{k}}\|_{2}}{\|\mathbf{x}\|_{2}},
\end{equation}
where $\mathbf{r}_{\mathbf{x},\hat{k}}$ is the generated perturbation, $\mathbf{x}$ is the input image, and $\mathbb{E}_{\mathbf{x}}$ is the expectation over the entire test data. Similar to the classification accuracy, the robustness of \textit{LS} is higher than \textit{CCN} for MNIST but lower for CIFAR10. \textit{CCN+LS} is the most robust model for both the datasets. 

\begin{table}[htp!]
\caption{Robustness of defense methods against DeepFool~\cite{moosavi2016deepfool} attack for MNIST~\cite{lecun2010mnist} (left) and CIFAR10~\cite{krizhevsky2009learning} (right). The best robustness is shown in bold and the second-best robustness is underlined}
\label{tbl:robustness}
\small
\hskip60pt
\begin{minipage}{.25\linewidth}
\begin{tabular}{|l|c|}
\hline
Method & Robustness \\
\hline\hline
SM & 0.225 \\
LS~\cite{warde201611} & \underline{0.571} \\
BReLU~\cite{zantedeschi2017efficient} & 0.493\\
\hline
L2SM & 0.441 \\
CCN & 0.479\\
CCN+LS & \textbf{0.827}\\
\hline
\end{tabular}
\end{minipage}
\hskip50pt
\begin{minipage}{.25\linewidth}
\begin{tabular}{|l|c|}
\hline
Method & Robustness \\
\hline\hline
SM & 0.014 \\
LS~\cite{warde201611} & 0.185 \\
BReLU~\cite{zantedeschi2017efficient} & 0.057\\
\hline
L2SM & 0.285 \\
CCN & \underline{0.549}\\
CCN+LS & \textbf{0.625}\\
\hline
\end{tabular}
\end{minipage}
\vspace*{-4mm}
\end{table}

\subsubsection{Adversarial Training}
\label{sec:adv_training}


Goodfellow et al.~\cite{goodfellow2014explaining} proposed to train the network simultaneously with original and crafted adversarial images to improve it's stability. However, adversarial training is sensitive to the value of $\epsilon$ that is used to train the network on. We train the models on real and adversarial images crafted from FGSM attack with $\epsilon = 0.3$ for MNIST, and $\epsilon = \frac{8}{255}$ for CIFAR10 dataset. The classification accuracies of adversarially trained models are reported in Table~\ref{tbl:wb_adv} for MNIST and CIFAR10 datasets. The results show that adversarial training improves the robustness for all the models. The performances of the models on MNIST dataset are comparable to each other with accuracy values above $95\%$. CIFAR10 results show better distinct between performance of various models. Most of the the models perform well for the $\epsilon$ value with which they were trained, but the performance degrades on other $\epsilon$ values. The models \textit{CCN} and \textit{CCN+LS} are are less sensitive to the training perturbation amount, and consistently stable across all the $\epsilon$ values.

\begin{table}[htp!]
\caption{Accuracy ($\%$) with adversarial training against FGSM~\cite{goodfellow2014explaining} attack on MNIST~\cite{lecun2010mnist} and CIFAR10~\cite{krizhevsky2009learning} testset. The best accuracy is shown in bold and the second-best accuracy is underlined}
\label{tbl:wb_adv}
\begin{center}
\tabcolsep=0.10cm
\scalebox{0.875}{
\begin{tabular}{|c||c|c|c|c|c||c|c|c|c|c|}
\hline
 & \multicolumn{5}{c|}{MNIST} & \multicolumn{5}{c|}{CIFAR10}\\
\hline
Method & $\epsilon$=0 & $\epsilon$=0.1 & $\epsilon$=0.2 & $\epsilon$=0.3 & $\epsilon$=0.4 & $\epsilon$=0 & $\epsilon$=2 & $\epsilon$=4 & $\epsilon$=8 & $\epsilon$=16\\
\hline\hline
SM & \textbf{99.49} & 96.51 & 98.67 & 99.15 & 96.06 & 83.8 & 71.56 & 67.18 & 71.46 & 43.09\\
LS~\cite{warde201611} & 99.45 & \underline{98.4} & 98.77 & 99.31 & \textbf{98.79} & 83.28 & \textbf{77.68} & 69.07 & 52.13 & 34.04\\
BReLU~\cite{zantedeschi2017efficient} & 99.43 & 96.79 & \textbf{98.93} & \textbf{99.42} & \underline{97.12} & 85.37 & 72.93 & 67.08 & \textbf{80.73} & 31.01\\
\hline
L2SM & 99.47 & 96.88 & 98.7 & \underline{99.32} & 84.03 & 84.8 & 73.08 & 68.31 & 74.95 & 49.64\\
CCN & 99.44 & 98.18 & 98.71 & 99.09 & 96.9 & \underline{87.08} & 73.27 & \underline{75.8} & 76.78 & \underline{61.97}\\
CCN+LS & \underline{99.48} & \textbf{98.65} & \underline{98.8} & 98.93 & 94.13 & \textbf{87.17} & \underline{76.68} & \textbf{77.53} & \underline{77.66} & \textbf{70.17}\\
\hline
\end{tabular}
}
\end{center}
\vspace*{-8mm}
\end{table}

\subsubsection{Evaluation on ImageNet dataset}
\label{sec:imagenet}

We also analyze the adversarial robustness of the proposed methods on ILSVRC~\cite{russakovsky2015imagenet} object classification dataset. The dataset contains $1.2M$ training images and $50k$ validation images with $1000$ class labels. The dataset is more challenging than MNIST and CIFAR10 as it contains large images in real-world settings. We use off-the-shelf VGG11~\cite{simonyan2014very} network for training the models. Table~\ref{tbl:imagenet} provides the classification accuracy of various defense methods on adversarial examples crafted using FGSM~\cite{goodfellow2014explaining} and DeepFool~\cite{moosavi2016deepfool} attacks. We use the $\epsilon$ values of \{$0.1$,$0.3$,$0.5$,$1.0$\} normalized by $255$ to craft the FGSM attack. From the table, we find that \textit{LS} and \textit{CCN+LS} are more robust to FGSM attack. On DeepFool~\cite{moosavi2016deepfool} attack, \textit{CCN+LS} is the most robust model followed by \textit{LS}.

Additionally, we see a significant improvement in top-1 accuracy($\%$) by using compact learning framework. \textit{L2SM} achieves a top-1 accuracy of $68.78\%$ while \textit{CCN} achieves the accuracy of $68\%$, which is $8\%$ gain over the baseline \textit{SM} model. Having large number of classes improves the discriminative capacity of compact learning leading to improved accuracy. We intend to analyze the effect of \textit{compactness} for large number of classes in future work.

\begin{table}[htp!]
\caption{Accuracy ($\%$) on ILSVRC~\cite{russakovsky2015imagenet} validation set for FGSM~\cite{goodfellow2014explaining} attack (left). Robustness of defense methods against DeepFool~\cite{moosavi2016deepfool} attack (right). The best accuracy is shown in bold and the second-best accuracy is underlined}
\label{tbl:imagenet}
\small
\begin{minipage}{.25\linewidth}
\begin{tabular}{|l|c|c|c|c|c|c|}
\hline
Method & $\epsilon$ = 0 & $\epsilon$ = 0.1 & $\epsilon$ = 0.3 & $\epsilon$ = 0.5 & $\epsilon$ = 1.0 \\
\hline\hline
SM & 60.52 & 47.19 & 25.72 & 17.02 & 10.14\\
LS~\cite{warde201611} & 63.80 & \textbf{58.84} & \textbf{43.17} & \textbf{32.55} & \underline{20.21}\\
BReLU~\cite{zantedeschi2017efficient} & 67.08 & 53.05 & 27.35 & 17.50 & 10.75\\
\hline
L2SM & \textbf{68.78} & \underline{56.66} & 33.74 & 22.95 & 13.92\\
CCN & \underline{68.00} & 51.43 & 28.57 & 19.54 & 12.68\\
CCN+LS & 66.64 & 53.22 & \underline{36.63} & \underline{29.90} & \textbf{23.53}\\
\hline
\end{tabular}
\end{minipage}
\hskip140pt
\begin{minipage}{.25\linewidth}
\begin{tabular}{|l|c|}
\hline
Method & Robustness (in $\%$) \\
\hline\hline
SM & 0.1325 \\
LS~\cite{warde201611} & \underline{0.2951} \\
BReLU~\cite{zantedeschi2017efficient} & 0.1463\\
\hline
L2SM & 0.1911 \\
CCN & 0.1439\\
CCN+LS & \textbf{0.4027}\\
\hline
\end{tabular}
\end{minipage}
\vspace*{-8mm}
\end{table}

\subsection{Black-Box Attacks}
\label{sec:bb_attack}

In a typical black-box attack, the attacker has no information about the network architecture, its parameters or the training dataset. The attacker can query the network and can get the output class label for a given input. We use the black-box attack proposed by Papernot~et~al.~\cite{papernot2017practical} to evaluate the proposed models. The model on which the attack has to be applied is the defense model, and the model learned to replicate the behavior of defense model is the substitute model. We treat our defense models as oracle, and train substitute models using LeNet~\cite{lecun1998gradient} architecture as described in~\cite{papernot2017practical}. Table~\ref{tbl:bb_mnist} reports the accuracy of the defense models against FGSM attack generated using the corresponding substitute models. We observe that \textit{CCN} and \textit{CCN+LS} models are most robust to practical black-box attacks, and are consistent across different $\epsilon$ values. Hence, the results show that the proposed models are robust to both white box as well as black box attacks.

\begin{table}[htp!]
\caption{Accuracy ($\%$) against practical black-box FGSM attack (with different $\epsilon$) on MNIST~\cite{lecun2010mnist} testset. The best accuracy is shown in bold and the second-best accuracy is underlined}
\label{tbl:bb_mnist}
\small
\begin{center}
\begin{tabular}{|l|c|c|c|c|c|c|}
\hline
Method & $\epsilon$ = 0.1 & $\epsilon$ = 0.2 & $\epsilon$ = 0.3 & $\epsilon$ = 0.4 \\
\hline\hline
SM & 98.49 & 87.44 & 55.41 & 32.84\\
LS~\cite{warde201611} & 98.89 & 91.12 & 52.53 & 29.28\\
BReLU~\cite{zantedeschi2017efficient} & 98.90 & 91.96 & 66.45 & \textbf{39.44}\\
\hline
L2SM & 98.99 & 93.37 & 56.67 & 26.63\\
CCN & \underline{99.15} & \underline{96.76} & \underline{77.27} & 28.19\\
CCN+LS & \textbf{99.24} & \textbf{97.33} & \textbf{84.14} & \underline{32.94}\\
\hline
\end{tabular}
\end{center}
\vspace*{-8mm}
\end{table}

\subsubsection{Transferability of Adversarial Samples}
\label{sec:transfer}
It has been shown in~\cite{papernot2017practical} that adversarial examples generated by one type of network can be used to fool a different type of network. This makes it easier for the attackers to generate adversarial samples using their independently trained models. Our defense model should be immune to the attacks generated by itself, as well to attacks generated from a different network.

Tables~\ref{tbl:transferability_mnist} and~\ref{tbl:transferability_cifar10} report the accuracies on transfered attacks between different defense models. ($^{*}$) in the tables indicates that the adversarial attacks were crafted and tested on the same network, causing maximum impact. We find that the networks are more vulnerable to the transfered attacks generated using the baseline model \textit{SM}. Among all models, \textit{CCN} and \textit{CCN+LS} are most robust to transferred attacks. Also, the attacks generated using these models are less likely to fool other models. This shows that \textit{CCN} and \textit{CCN+LS} provide a two-way defense. Firstly, these models would be less vulnerable to any unknown adversarial attacks. Secondly, the attacks generated using these models would be less harmful for any unknown network.

\begin{table}[htp!]
\caption{Accuracy ($\%$) on MNIST~\cite{lecun2010mnist} against transfer attacks crafted using FGSM ($\epsilon = 0.3$). The best accuracy is shown in bold and the second-best accuracy is underlined}
\label{tbl:transferability_mnist}
\small
\begin{center}
\begin{tabular}{|c||c|c|c|c|c|c|}
\hline
 & \multicolumn{6}{c|}{Attack crafted on}\\
\hline
Attack tested on & SM & LS~\cite{warde201611} & BReLU~\cite{zantedeschi2017efficient} & L2SM & CCN & CCN+LS\\
\hline\hline
SM & 31.76$^{*}$ & 61.41 & 54.22 & 77.88 & \underline{91.22} & \underline{93.84} \\
LS~\cite{warde201611} & 49.48 & 44.25$^{*}$ & 53.63 & 80.47 & \textbf{91.76} & \textbf{94.69} \\
BReLU~\cite{zantedeschi2017efficient} & 52.62 & 66.43 & 42.51$^{*}$ & 79.69 & 90.82 & 93.77 \\
\hline
L2SM & 52.01 & 64.93 & 58.91 & 60.32$^{*}$ & 90.65 & 93.59 \\
CCN & \underline{64.15} & \underline{75.7} & \underline{66.55} & \underline{82.7} & 81.38$^{*}$ & 93.53\\
CCN+LS & \textbf{70.18} & \textbf{80.83} & \textbf{71.49} & \textbf{85.29} & 90.49 & 89.73$^{*}$\\
\hline
\end{tabular}
\end{center}
\vspace*{-8mm}
\end{table}

\begin{table}[htp!]
\caption{Accuracy ($\%$) on CIFAR10~\cite{krizhevsky2009learning} against transfer attacks crafted using FGSM ($\epsilon = \frac{8}{255}$). The best accuracy is shown in bold and the second-best accuracy is underlined}
\label{tbl:transferability_cifar10}
\small
\begin{center}
\begin{tabular}{|c||c|c|c|c|c|c|}
\hline
 & \multicolumn{6}{c|}{Attack crafted on}\\
\hline
Attack tested on & SM & LS~\cite{warde201611} & BReLU~\cite{zantedeschi2017efficient} & L2SM & CCN & CCN+LS\\
\hline\hline
SM & 14.30$^{*}$ & 59.14 & 33.63 & 67.13 & 77.77 & 82.91 \\
LS~\cite{warde201611} & 31.16 & 54.07$^{*}$ & 33.28 & 67.26 & \underline{77.84} & \textbf{83.11} \\
BReLU~\cite{zantedeschi2017efficient} & 33.63 & 59.65 & 18.01$^{*}$ & 67.96 & \textbf{77.85} & \underline{82.95} \\
\hline
L2SM & 32.59 & 59.95 & 35.01 & 63.71$^{*}$ & 77.76 & 82.78 \\
CCN & \underline{41.33} & \textbf{63.6} & \textbf{42.37} & \underline{69.63} & 69.65$^{*}$ & 80.53 \\
CCN+LS & \textbf{41.48} & \underline{63.48} & \underline{42.29} & \textbf{69.98} & 73.74 & 76.8$^{*}$ \\
\hline
\end{tabular}
\end{center}
\vspace*{-8mm}
\end{table}

\subsection{Effect of Feature Normalization Methods}
\label{sec:norm}

In this experiment, we analyze the effect of Local Response Normalization (LRN)~\cite{krizhevsky2012imagenet} and batch-normalization~\cite{ioffe2015batch} on the robustness of the network and compare it to the proposed approach. Although both these methods normalizes the features based on local activations or input batch statistics, they do not ensure \textit{compactness}. LRN layer implements a form of lateral inhibition given by~(\ref{eq:lrn})

\begin{equation}
\small
\label{eq:lrn}
b_{x,y}^{i} = a_{x,y}^{i} / \left( k + \alpha \sum_{j=\max(0,i-n/2)}^{\min(N-1,i+n/2)} (a_{x,y}^{j})^{2} \right)^{\beta},
\end{equation}

where $a_{x,y}^{i}$ is the input activation at location $(x,y)$ and channel $i$, $b_{x,y}^{i}$ is the corresponding output activation. The values of constants $k$, $n$, $\alpha$ and $\beta$ are are set as provided in~\cite{krizhevsky2012imagenet}. On the other hand, batch-normalization~\cite{ioffe2015batch} reduces the \textit{internal covariate shift}, by normalizing the features across the input batch. We compare the performance of the models trained with \textit{LRN} and batch-normalization (\textit{BN}) with the baseline model \textit{SM} in Table~\ref{tbl:wb_lrn_bn}. We find from our experiments on MNIST and CIFAR-10 datasets that LRN or batch-normalization either improves marginally or weakens the robustness of the network. Our proposed models \textit{CCN} and \textit{CCN+LS} significantly outperforms \textit{LRN} and \textit{BN} models for both MNIST~\cite{lecun2010mnist} and CIFAR10~\cite{krizhevsky2009learning} datasets.

\begin{table}[htp!]
\small
\caption{Accuracy ($\%$) on MNIST~\cite{lecun2010mnist} and CIFAR10~\cite{krizhevsky2009learning} for FGSM~\cite{goodfellow2014explaining} attack. The best accuracy is shown in bold and the second-best accuracy is underlined}
\label{tbl:wb_lrn_bn}
\begin{center}
\tabcolsep=0.10cm
\scalebox{0.875}{
\begin{tabular}{|c||c|c|c|c|c||c|c|c|c|c|}
\hline
 & \multicolumn{5}{c|}{MNIST} & \multicolumn{5}{c|}{CIFAR10}\\
\hline
Method & $\epsilon$=0 & $\epsilon$=0.1 & $\epsilon$=0.2 & $\epsilon$=0.3 & $\epsilon$=0.4 & $\epsilon$=0 & $\epsilon$=2 & $\epsilon$=4 & $\epsilon$=8 & $\epsilon$=16\\
\hline\hline
SM & 99.50 & 92.61 & 63.71 & 31.76 & 19.39 & 91.14 & 60.82 & 34.62 & 14.30 & 8.57\\
BN~\cite{ioffe2015batch} & \textbf{99.58} & 88.48 & 30.72 & 11.20 & 9.24 & \textbf{92.34} & 59.47 & 40.63 & 26.7 & 16.58\\
LRN~\cite{krizhevsky2012imagenet} & 99.28 & 88.59 & 52.52 & 25.61 & 13.21 & \underline{91.33} & 60.68 & 34.82 & 14.99 & 9.08\\
\hline
CCN & \underline{99.50} & \underline{96.99} & \underline{91.64} & \underline{81.38} & \underline{60.65} & 90.54 & \underline{73.72} & \underline{71.47} & \underline{69.65} & \underline{66.23}\\
CCN+LS & 99.43 & \textbf{97.26} & \textbf{93.68} & \textbf{89.73} & \textbf{75.95} & 90.51 & \textbf{80.93} & \textbf{79.2} & \textbf{76.8} & \textbf{71.38}\\
\hline
\end{tabular}
}
\end{center}
\vspace*{-8mm}
\end{table}

%% file: supplementary.tex
\section{Network Architectures}
\label{sec:arch}

Table~\ref{tbl:arch} provides the network architectures used for training. For training on MNIST~\cite{lecun2010mnist}, we use the architecture proposed by Papernot~et~al.~\cite{papernot2016distillation}. For training on CIFAR10~\cite{krizhevsky2009learning}, we use the standard VGG11~\cite{simonyan2014very} network.

\begin{table}[htp!]
\small
\caption{Overview of network architectures for MNIST~\cite{lecun2010mnist} and CIFAR10~\cite{krizhevsky2009learning} datasets}
\label{tbl:arch}
\begin{center}
\begin{tabular}{|l|c|c|}
\hline
Layer & MNIST & CIFAR10 \\
\hline\hline
Conv+ReLU & 32 filters (3x3) & 64 filters (3x3)\\
\hline
Conv+ReLU & 32 filters (3x3) & 128 filters (3x3)\\
\hline
MaxPool & 2x2 & 2x2\\
\hline
Conv+ReLU & 64 filters (3x3) & 256 filters (3x3)\\
\hline
Conv+ReLU & 64 filters (3x3) & 256 filters (3x3)\\
\hline
MaxPool & 2x2 & 2x2\\
\hline
Conv+ReLU & - & 512 filters (3x3)\\
\hline
Conv+ReLU & - & 512 filters (3x3)\\
\hline
MaxPool & - & 2x2\\
\hline
Conv+ReLU & - & 512 filters (3x3)\\
\hline
Conv+ReLU & - & 512 filters (3x3)\\
\hline
MaxPool & - & 2x2\\
\hline
FC+ReLU & 200 units & 512 units\\
\hline
FC+ReLU & 200 units & 512 units\\
\hline
Softmax & 10 units & 10 units\\
\hline
\end{tabular}
\end{center}
\vspace*{-8mm}
\end{table}

\section{C\&W Attack}
We evaluate different network defense methods against the attack proposed by Carlini and Wagner~\cite{carlini2017towards} (C\&W Attack). $L_{2}$-distance metric of the adversarial perturbation is optimized to generate the attack, as given by:

\begin{equation}
\small
\label{eq:cw}
minimize~\|\frac{1}{2}(\tanh(\mathbf{w})+1)-\mathbf{x}\|_{2}^{2} + c~f(\frac{1}{2}(\tanh(\mathbf{w})+1)),
\end{equation}
where $\mathbf{x}$ is the input image, and $\mathbf{w}$ is the variable to be optimized. For the target class $t$, the function $f$ corresponding to input $\tilde{\mathbf{x}}$ is given by:

\begin{equation}
\small
\label{eq:cwf}
f(\tilde{\mathbf{x}}) = \max (\max\{Z(\tilde{\mathbf{x}})_{i}:i \neq t\}-Z(\tilde{\mathbf{x}})_{t},-\kappa),
\end{equation}

where $Z(\tilde{\mathbf{x}})_{i}$ is the logit value for the $i^{th}$ class, and $\kappa$ is the confidence with which the misclassification occurs.  In our experiments, we set as the target class $t$ to the class with second-highest classification score. The confidence $\kappa$ is set to zero. The maximum number of iterations is set to $1000$, and the search step is limited to $6$. We observe that C\&W attack is approximately $100\times$ slower than FGSM attack~\cite{goodfellow2014explaining}.

\begin{table}[htp!]
\caption{Mean $L_{2}$-distance between the input and perturbed image on MNIST dataset, along with the success probability of generating the C$\&$W attack. The best result is shown in bold and the second-best result is underlined}
\label{tbl:cw_mnist}
\small
\begin{center}
\begin{tabular}{|l|c|c|}
\hline
Method & Mean Distance & Success Prob.\\
\hline\hline
SM & 1.393 & 100\\
LS~\cite{warde201611} & \underline{1.626} & 100\\
BReLU~\cite{zantedeschi2017efficient} & 1.473 & 100\\
\hline
L2SM & 1.453 & 100\\
CCN & 1.449 & 100\\
CCN+LS & \textbf{2.079} & \textbf{96.54}\\
\hline
\end{tabular}
\end{center}
\end{table}

\begin{table}[htp!]
\caption{Mean $L_{2}$-distance between the input and perturbed image on CIFAR10 dataset, along with the success probability of generating the C$\&$W attack. The best result is shown in bold and the second-best result is underlined}
\label{tbl:cw_cifar10}
\small
\begin{center}
\begin{tabular}{|l|c|c|}
\hline
Method & Mean Distance & Success Prob.\\
\hline\hline
SM & 0.277 & 100\\
LS~\cite{warde201611} & 0.413 & 100\\
BReLU~\cite{zantedeschi2017efficient} & 0.312 & 99.13\\
\hline
L2SM & \textbf{0.717} & \textbf{94.26}\\
CCN & 0.352 & 99.95\\
CCN+LS & \underline{0.471} & \underline{99.08}\\
\hline
\end{tabular}
\end{center}
\end{table}

Tables~\ref{tbl:cw_mnist} and \ref{tbl:cw_cifar10} show the performance of difference defense methods against C\&W attack~\cite{carlini2017towards}, on MNIST and CIFAR10 datasets respectively. We report the mean $L_{2}$-distance between the original and the adversarial samples, along with the success probability of generating the C\&W attack. A higher mean distance signifies that the defense method is more robust. For MNIST dataset, we observe a behavior similar to the DeepFool~\cite{moosavi2016deepfool} attack. The \textit{CCN+LS} model significantly outperforms the other defense approaches, followed by \textit{LS} model.
For CIFAR10 dataset, we find that $L_{2}$-softmax loss (\textit{L2SM}) achieves the best performance with a mean distance of $0.717$. It is followed by \textit{CCN+LS} with a mean distance of $0.471$.


\section{Ablation Study}
\label{sec:ablation}

\subsection{Robustness analysis of \textit{compactness} for each layer}

In this section, we analyze the layer-wise effect of \textit{compactness} on the robustness of a network. We replace one layer from a CNN with the proposed compact convolutional module, and train the entire network on MNIST~\cite{lecun2010mnist} dataset. We use the same network architecture as given in Table~\ref{tbl:arch}. 

Table~\ref{tbl:layer_ccn} provides the accuracy of different networks, where only one layer is implemented using compact convolution, against FGSM~\cite{goodfellow2014explaining} attack ($\epsilon$ = 0.3) on MNIST~\cite{lecun2010mnist} testset. We see that the effect of \textit{compactness} on the network's robustness decreases with the depth of the network till a certain layer, after which it increases consistently. The network without any compact layer is equivalent to the \textit{SM} model, and achieves the accuracy of $31.76\%$. Applying compact module only to the \textit{conv1} layer improves the accuracy to $44.38\%$. Each of the \textit{conv2}, \textit{conv3} and \textit{conv4} layers decreases the accuracy marginally. The behavior gets reversed in fully connected layers, for which the accuracy increases when compact module is applied to a deeper layer. Applying compact module only to the last fully connected layer (fc3) is equivalent to the \textit{L2SM} model, and achieves the best accuracy of $60.32\%$. This shows that \textit{compactness} has moderate effect in the shallower layers of the network, and has maximum effect in the deepest layer of the network.

\begin{table}[htp!]
\caption{Accuracy ($\%$) against  FGSM attack (with $\epsilon$ = 0 and $\epsilon$ = 0.3) on MNIST~\cite{lecun2010mnist} testset. The best accuracy is shown in bold and the second-best accuracy is underlined}
\label{tbl:layer_ccn}
\begin{center}
\begin{tabular}{|l|c|c|}
\hline
Compact Layer & $\epsilon$ = 0 & $\epsilon$ = 0.3\\
\hline\hline
None & 99.50 & 31.76\\
\hline
conv1 & 99.50 & 44.38\\
\hline
conv2 & 99.51 & 42.34\\
\hline
conv3 & 99.50 & 39.80\\
\hline
conv4 & 99.54 & 39.04\\
\hline
fc1 & \textbf{99.58} & 44.19\\
\hline
fc2 & \underline{99.57} & \underline{58.71}\\
\hline
fc3 & 99.48 & \textbf{60.32}\\
\hline
\end{tabular}
\end{center}
\end{table}

\subsection{t-SNE visualization of adversarial features}

\begin{figure}[t]
	\begin{center}
		\includegraphics[width=12cm, height=9cm]{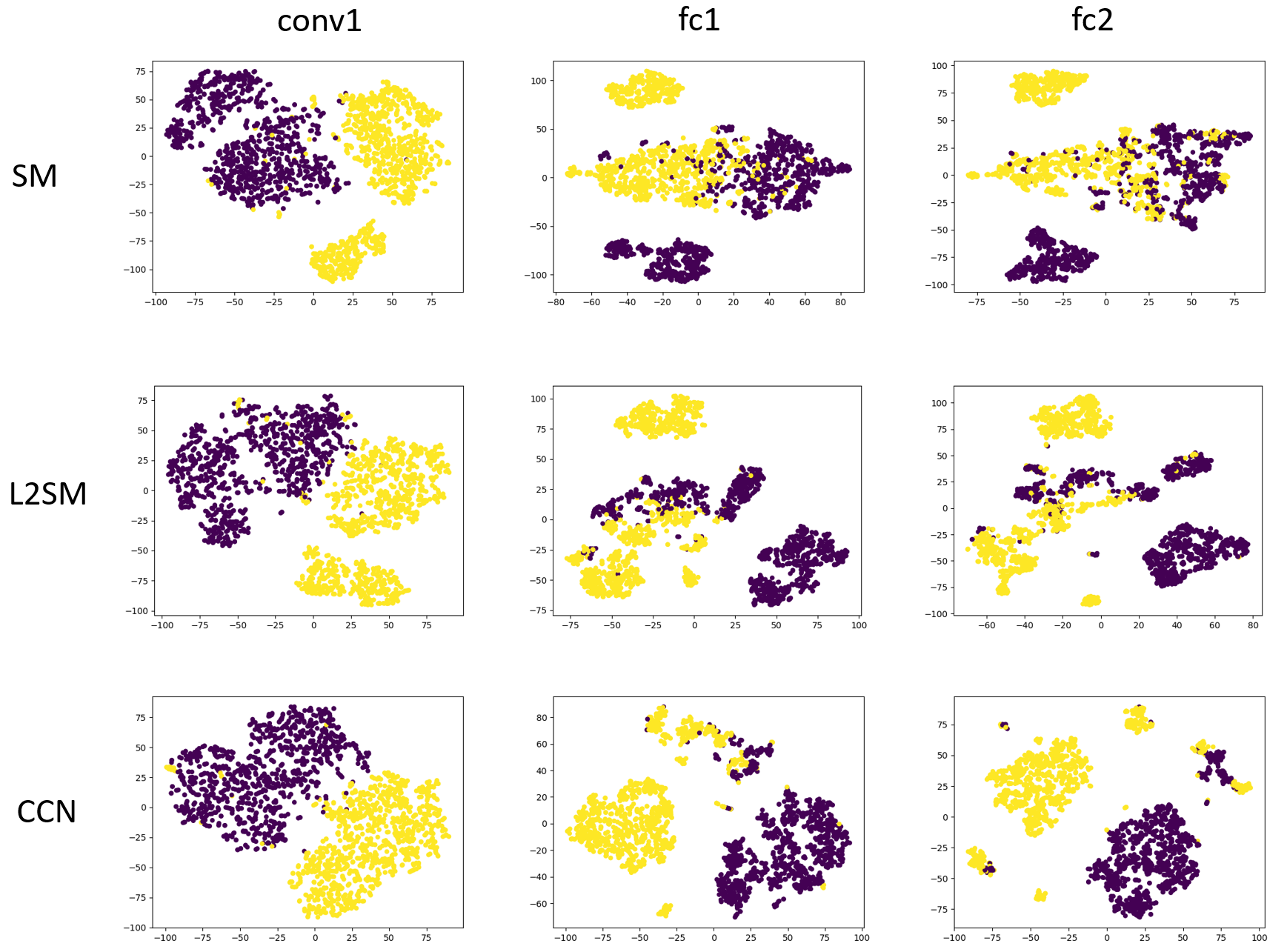}
\end{center}
\caption{t-SNE plots for the features of \textit{conv1}, \textit{fc1} and \textit{fc2} for models \textit{SM}, \textit{L2SM} and \textit{CCN} respectively. The features are computed for the adversarial images generated using FGSM attack ($\epsilon$ = 0.3), for two randomly chosen classes from MNIST~\cite{lecun2010mnist}}
\label{fig:tsne}
\end{figure}

In this subsection, we analyze the effect of \textit{compactness} on the separability of features. Fig.~\ref{fig:tsne} provides a two dimensional t-SNE~\cite{maaten2008visualizing} visualization of the features obtained after \textit{conv1}, \textit{fc1} and \textit{fc2} layers of the network, for \textit{SM}, \textit{L2SM} and \textit{CCN} models respectively. The features are computed for the adversarial images generated using FGSM attack ($\epsilon$ = 0.3), for two randomly chosen classes from MNIST~\cite{lecun2010mnist}. From the figure, it is evident that the features from \textit{CCN} and \textit{L2SM} are more separable compared to features from \textit{SM}. The difference in separability is minimum for the initial layers of the network (such as conv1), and dominant for the deeper layers (such as fc1 and fc2).

\subsection{Analysis of learned network filters}

\begin{figure}[htp!]
	\begin{center}
		\includegraphics[width=12cm, height=7cm]{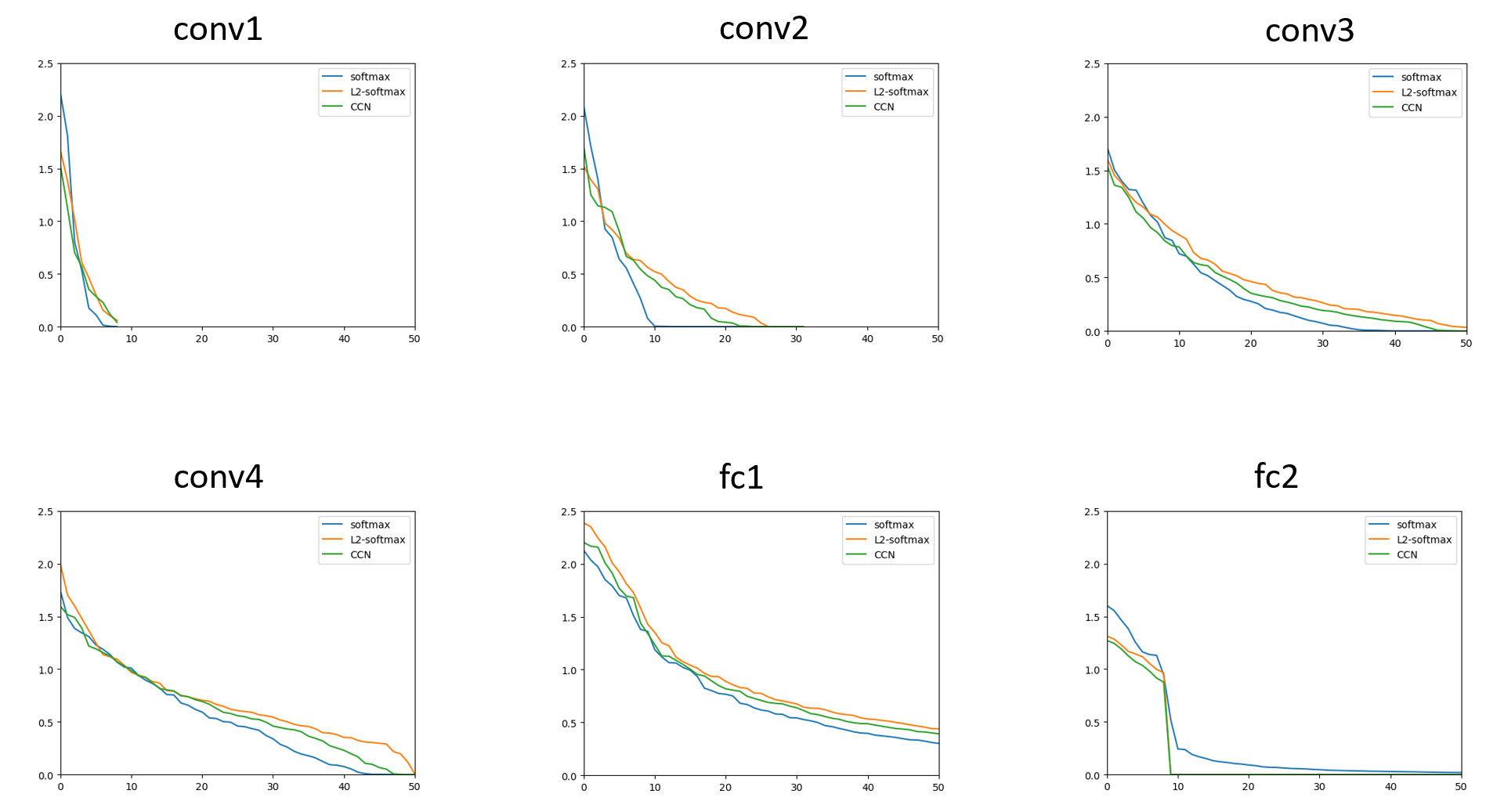}
\end{center}
\caption{Average singular value (SV) spectrum showing top 50 SVs of the weight matrix for each layer of the network.}
\label{fig:jacobian}
\end{figure}

In this subsection, we analyze and compare the characteristics of the weight filters learned for \textit{SM}, \textit{L2SM} and \textit{CCN} models. We compute the top 50 singular values (SVs) for the weight matrix of \textit{conv1}, \textit{conv2}, \textit{conv3}, \textit{conv4}, \textit{fc1} and \textit{fc2} layers, and plot their magnitudes in Fig.~\ref{fig:jacobian}. From the figure, we observe that the \textit{fc2} layer of \textit{CCN} and \textit{L2SM} models has fewer dominant singular values that decay very rapidly compared to SVs of \textit{SM} model. This suggests that \textit{CCN} and \textit{L2SM} models have a strong suppression for the trailing dimensions, which makes them stable to the adversarial variations in the data. For the initial layers of the networks, the SVs for \textit{CCN}, \textit{L2SM} and \textit{SM} models are dominant throughout, since they are less invariant to input deformations.

%% file: conclusion.tex
\section{Conclusion}
\label{sec:conclusion}

In this paper, we show that learning features by imposing \textit{compactness} constraints can improve a network's robustness against adversarial attacks. The $L_{2}$-Softmax Loss, that ensures feature \textit{compactness}, provide better robustness compared to naive softmax loss. This property is applied to each layer of the network using compact convolutional modules (CCN), which significantly reduces the network's vulnerability to adversarial perturbations. In future, we would further analyze the necessary properties for a network to be robust, and build sophisticated architectures that are provably robust to adversarial attacks.

%% file: main_iarpa.bbl
\begin{thebibliography}{10}

\bibitem{szegedy2013intriguing}
Szegedy, C., Zaremba, W., Sutskever, I., Bruna, J., Erhan, D., Goodfellow, I.,
  Fergus, R.:
\newblock Intriguing properties of neural networks.
\newblock arXiv preprint arXiv:1312.6199 (2013)

\bibitem{goodfellow2014explaining}
Goodfellow, I.J., Shlens, J., Szegedy, C.:
\newblock Explaining and harnessing adversarial examples.
\newblock arXiv preprint arXiv:1412.6572 (2014)

\bibitem{moosavi2016deepfool}
Moosavi-Dezfooli, S.M., Fawzi, A., Frossard, P.:
\newblock Deepfool: a simple and accurate method to fool deep neural networks.
\newblock In: Proceedings of the IEEE Conference on Computer Vision and Pattern
  Recognition. (2016)  2574--2582

\bibitem{moosavi2016universal}
Moosavi-Dezfooli, S.M., Fawzi, A., Fawzi, O., Frossard, P.:
\newblock Universal adversarial perturbations.
\newblock arXiv preprint arXiv:1610.08401 (2016)

\bibitem{krizhevsky2009learning}
Krizhevsky, A., Hinton, G.:
\newblock Learning multiple layers of features from tiny images.
\newblock (2009)

\bibitem{sankaranarayanan2017regularizing}
Sankaranarayanan, S., Jain, A., Chellappa, R., Lim, S.N.:
\newblock Regularizing deep networks using efficient layerwise adversarial
  training.
\newblock arXiv preprint arXiv:1705.07819 (2017)

\bibitem{tramer2017ensemble}
Tram{\`e}r, F., Kurakin, A., Papernot, N., Boneh, D., McDaniel, P.:
\newblock Ensemble adversarial training: Attacks and defenses.
\newblock arXiv preprint arXiv:1705.07204 (2017)

\bibitem{ranjan2017l2}
Ranjan, R., Castillo, C.D., Chellappa, R.:
\newblock L2-constrained softmax loss for discriminative face verification.
\newblock arXiv preprint arXiv:1703.09507 (2017)

\bibitem{papernot2016limitations}
Papernot, N., McDaniel, P., Jha, S., Fredrikson, M., Celik, Z.B., Swami, A.:
\newblock The limitations of deep learning in adversarial settings.
\newblock In: Security and Privacy (EuroS\&P), 2016 IEEE European Symposium on,
  IEEE (2016)  372--387

\bibitem{carlini2017towards}
Carlini, N., Wagner, D.:
\newblock Towards evaluating the robustness of neural networks.
\newblock In: Security and Privacy (SP), 2017 IEEE Symposium on, IEEE (2017)
  39--57

\bibitem{lu2017safetynet}
Lu, J., Issaranon, T., Forsyth, D.:
\newblock Safetynet: Detecting and rejecting adversarial examples robustly.
\newblock arXiv preprint arXiv:1704.00103 (2017)

\bibitem{metzen2017detecting}
Metzen, J.H., Genewein, T., Fischer, V., Bischoff, B.:
\newblock On detecting adversarial perturbations.
\newblock arXiv preprint arXiv:1702.04267 (2017)

\bibitem{grosse2017statistical}
Grosse, K., Manoharan, P., Papernot, N., Backes, M., McDaniel, P.:
\newblock On the (statistical) detection of adversarial examples.
\newblock arXiv preprint arXiv:1702.06280 (2017)

\bibitem{feinman2017detecting}
Feinman, R., Curtin, R.R., Shintre, S., Gardner, A.B.:
\newblock Detecting adversarial samples from artifacts.
\newblock arXiv preprint arXiv:1703.00410 (2017)

\bibitem{carlini2017adversarial}
Carlini, N., Wagner, D.:
\newblock Adversarial examples are not easily detected: Bypassing ten detection
  methods.
\newblock arXiv preprint arXiv:1705.07263 (2017)

\bibitem{miyato2017virtual}
Miyato, T., Maeda, S.i., Koyama, M., Ishii, S.:
\newblock Virtual adversarial training: a regularization method for supervised
  and semi-supervised learning.
\newblock arXiv preprint arXiv:1704.03976 (2017)

\bibitem{shaham2015understanding}
Shaham, U., Yamada, Y., Negahban, S.:
\newblock Understanding adversarial training: Increasing local stability of
  neural nets through robust optimization.
\newblock arXiv preprint arXiv:1511.05432 (2015)

\bibitem{kurakin2016adversarial}
Kurakin, A., Goodfellow, I., Bengio, S.:
\newblock Adversarial machine learning at scale.
\newblock arXiv preprint arXiv:1611.01236 (2016)

\bibitem{dziugaite2016study}
Dziugaite, G.K., Ghahramani, Z., Roy, D.M.:
\newblock A study of the effect of jpg compression on adversarial images.
\newblock arXiv preprint arXiv:1608.00853 (2016)

\bibitem{guo2017countering}
Guo, C., Rana, M., Cisse, M., van~der Maaten, L.:
\newblock Countering adversarial images using input transformations.
\newblock arXiv preprint arXiv:1711.00117 (2017)

\bibitem{gu2014towards}
Gu, S., Rigazio, L.:
\newblock Towards deep neural network architectures robust to adversarial
  examples.
\newblock arXiv preprint arXiv:1412.5068 (2014)

\bibitem{cisse2017parseval}
Cisse, M., Bojanowski, P., Grave, E., Dauphin, Y., Usunier, N.:
\newblock Parseval networks: Improving robustness to adversarial examples.
\newblock In: International Conference on Machine Learning. (2017)  854--863

\bibitem{papernot2016distillation}
Papernot, N., McDaniel, P., Wu, X., Jha, S., Swami, A.:
\newblock Distillation as a defense to adversarial perturbations against deep
  neural networks.
\newblock In: Security and Privacy (SP), 2016 IEEE Symposium on, IEEE (2016)
  582--597

\bibitem{warde201611}
Warde-Farley, D., Goodfellow, I.:
\newblock Adversarial perturbations of deep neural networks.
\newblock Perturbations, Optimization, and Statistics (2016)  311

\bibitem{zantedeschi2017efficient}
Zantedeschi, V., Nicolae, M.I., Rawat, A.:
\newblock Efficient defenses against adversarial attacks.
\newblock arXiv preprint arXiv:1707.06728 (2017)

\bibitem{lecun2010mnist}
LeCun, Y., Cortes, C., Burges, C.J.:
\newblock \textsc{MNIST} handwritten digit database.
\newblock AT\&T Labs [Online]. Available: http://yann. lecun. com/exdb/mnist
  \textbf{2} (2010)

\bibitem{russakovsky2015imagenet}
Russakovsky, O., Deng, J., Su, H., Krause, J., Satheesh, S., Ma, S., Huang, Z.,
  Karpathy, A., Khosla, A., Bernstein, M.,  et~al.:
\newblock Imagenet large scale visual recognition challenge.
\newblock International Journal of Computer Vision \textbf{115}(3) (2015)
  211--252

\bibitem{rauber2017foolbox}
Rauber, J., Brendel, W., Bethge, M.:
\newblock Foolbox v0. 8.0: A python toolbox to benchmark the robustness of
  machine learning models.
\newblock arXiv preprint arXiv:1707.04131 (2017)

\bibitem{simonyan2014very}
Simonyan, K., Zisserman, A.:
\newblock Very deep convolutional networks for large-scale image recognition.
\newblock arXiv preprint arXiv:1409.1556 (2014)

\bibitem{krizhevsky2012imagenet}
Krizhevsky, A., Sutskever, I., Hinton, G.E.:
\newblock Imagenet classification with deep convolutional neural networks.
\newblock In: Advances in neural information processing systems. (2012)
  1097--1105

\bibitem{ioffe2015batch}
Ioffe, S., Szegedy, C.:
\newblock Batch normalization: Accelerating deep network training by reducing
  internal covariate shift.
\newblock In: International Conference on Machine Learning. (2015)  448--456

\bibitem{papernot2017practical}
Papernot, N., McDaniel, P., Goodfellow, I., Jha, S., Celik, Z.B., Swami, A.:
\newblock Practical black-box attacks against machine learning.
\newblock In: Proceedings of the 2017 ACM on Asia Conference on Computer and
  Communications Security, ACM (2017)  506--519

\bibitem{lecun1998gradient}
LeCun, Y., Bottou, L., Bengio, Y., Haffner, P.:
\newblock Gradient-based learning applied to document recognition.
\newblock Proceedings of the IEEE \textbf{86}(11) (1998)  2278--2324

\bibitem{maaten2008visualizing}
Maaten, L.v.d., Hinton, G.:
\newblock Visualizing data using t-sne.
\newblock Journal of machine learning research \textbf{9}(Nov) (2008)
  2579--2605

\end{thebibliography}
